# Chatbot System Architecture

Moataz Mohammed[*1] and Mostafa M. Aref[**2]

[*]*Computer Science Department, Faculty of computer and Information Sciences, Ain-shams University*

[**]*Computer Science Department, Faculty of Computer and Information Sciences, Ain-shams University*

[1]`moataz.mohammed@cis.asu.edu.eg`
[2]`aref_99@yahoo.com`

*Abstract*— The conversational agents is one of the most interested topics in computer science field in the recent decade. Which can be composite from more than one subject in this field, which you need to apply Natural Language Processing Concepts and some Artificial Intelligence Techniques such as Deep Learning methods to make decision about how should be the response. This paper is dedicated to discuss the system architecture for the conversational agent and explain each component in details.

*Keywords:* Conversational Agents, Chatbots, System agent, Dialog System, Natural Language Understanding, Natural Language Processing, Deep Learning, System Architecture.

## 1 INTRODUCTION

A chatbot (conversational agent (CA), dialogue system) is a computer software that acts as an interface between human users and a software application, using spoken or written natural language as the primary means of communication. Examples on this Apple Siri, Google Assistant and Amazon Alexa. In the recent decade most of companies throughout the world use it as a service to improve the way of communicating with clients in any time and the response is instantly which there is no late which this provide to the client the comfortability. So this is so important technology service should be improved continuously. In this paper we will discuss the CA system architecture. There are many variant system architectures which the developers can follow them to develop the chatbot(conversational agent) but any one of them mush have three significant core components: Natural Language Understanding(NLU) is responsible about understanding the user's utterances meaning and put them in representational format, then The Dialog Manager(DM) which it's the most important component about acting as a mediator which receive the representational format from the NLU and processing them then send the responses for the Natural Language Generator(NLG) which is the last core component in any Conversational Agent(CA) architecture. It takes the responses from the Dialogue Manager (DM) and check if there are more than one valid response taking the one with highest priority. Finally produce the response in the final format which may be text or Speech.

Social chatbots' appeal lies not only in their ability to respond to users' diverse requests, but also in being able to establish an emotional connection with users. The latter is done by satisfying users' need for communication. How to make the social chatbot more Human Like about Emotional Quotient (EQ) **[1]**. Also it can make an examination on the influence of its responsiveness and embodiment on the answers people give in response to sensitive and non-sensitive questions **[2]**. AI and deep learning can help us in building such chatbots that improve the lives of people who have busy schedule to easily keep a check on their health **[3]**. The access to large-scale data and real-world feedback can drive faster progress in research **[4]**.

## 2 LITERATURE REVIEW

When we made a deep reading to a survey papers, an articles and journals. We found many conversational agents each one with its own architecture. Like Amazon Alexa bot which take the data of the user that may be voice and make an Automatic Speech Recognition ASR about the Amazon ASR service. Then make a processing to the received data about some Amazon Web Services AWS and using Amazon DynamoDB to store the conversations and its state**[4]**. Google Assistant the most successful bot at all.Which receive the recordings from users, then sending these recordings to google's servers which works on making processing. It makes break down the voices into individual sounds then try to match every single sound with the most similar word's pronunciation one which is stored on google database. Then it identifies the required task through some matched key words. And other more agents so we mentioned just two examples. From all these different architectures we will talk in this paper about the most commonly generic one which include three main components Natural Language Understanding, Dialog Manager, and Natural Language Generator.

## 3 SYSTEM ARCHITECTURE

In this section we will discuss each component in the architecture in details which we will talk about the three core components of any chatbot (Conversational Agent) and it's subs, First Component is the Natural Language Understanding(NLU) and its role in the system. After this we will talk about the subcomponents of the NLU such as Topic Detection, Intent Analysis, and Entity Linking. Second Component is the Dialog Manager(DM) and its subs such as Rule-base, Knowledge-base, Neural Network Reply Generation and the Online Information Retrieval. Finally the Natural Language Generation(NLG) or Reply Generator and it subs like Content Filter and Engagement Ranking. the following figure1 **[5]** can be considered as a simulation for these components.

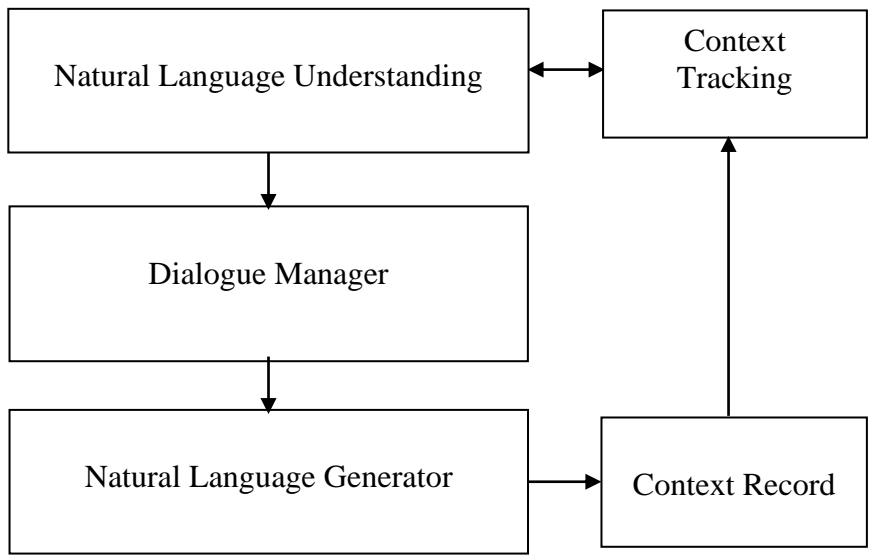

Figure 1: System Architecture of the Conversational Agent[e.g. chatbot]

*A. Natural Language Understanding*

Natural language understanding (NLU) is the first core component of the conversational agents which is responding about providing a semantic representation for user utterance **[6]** such as an in form of logic or class's intent, extracting the "meaning" of an utterance **[7]**. Parsing is the main task of - an NLU, which take the string of words and providing a linguistic structure for the utterance. Implementation-dependent is the method which an NLU uses it to parse the input and can utilize context free grammars, pattern matching, or data-driven approaches as we see in figure2. NLU outputs have to able to be tackled by a dialogue manager **[8]**.

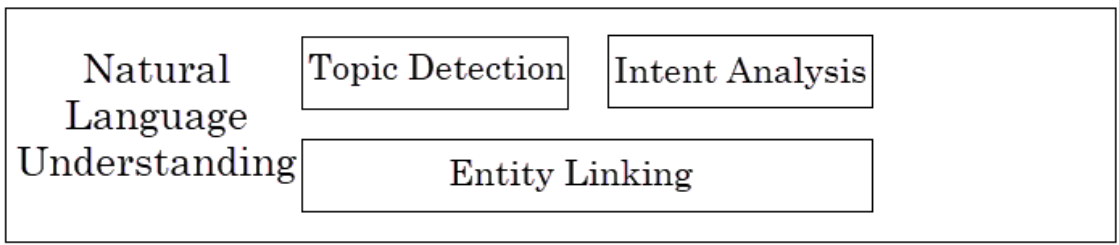

Figure 2: Natural Language Understanding [First Core Component]

1) *Topic Detection: from figure2. Topic Detection is One of the most important steps in developing the conversational agents NLU stage is to identify the Topic of the text (Conversation) for natural free-form interactions. Accurate tracking of the conversation topic can be a valuable signal for a system for dialog generation* **[9]**. *Examples on techniques can be used for topic classification are DANs and ADANs.*

2) *Intent Analysis: An intent is a group of utterances with similar meaning. See the following example of these two sentences: "I want to make a reservation in an Italian restaurant" and "I need a table in a pizzeria". They both have the same meaning. That means the intent analysis goal is to identify the similarities between words in meaning. If you can't understand a user, your bot will be useless whatever the effort you consume to develop the other components such as dialogue management. How we can Teaching semantics to a machine. We can do this about word vectors algorithms such as Word2Vec or Glove.* **[10]**

3) *Entity Linking: the bottom component in figure2 consists of a Disambiguation model and Named Entity Recognition(NER) and a template selection model. NER* **[11]** *links entity mentioned in a text to a dictionary or knowledge base, local or remote such as the whole web, to make sense of an entity and know what it is. This is a significant step for allowing the chatbot to understand conversation topics and generate appropriate responses. As it links the words of the user's utterances with concepts and subtexts in the real word. Options are to use StanfordNLP, TAGME* **[12]** *or web mining via a search engine API. TAGME can take the input text and returns a list of entities with their concept titles from wikipedia, which in turn can be converted to nodes in the Wikidata knowledge graph.*

B. *Context Tracking*

When a sentence from a user appears, the chatbot obtains the most recent utterances of that user from the chat history database. The Stanford CoreNLP toolkit **[13]** can be used to resolve coreference. if a coreferent was identified,The pronouns and the mentions of entities in the new sentence will be replaced.

C. *Dialogue Manager*

Dialogue Manager is The second core component in any chatbot and we can differentiate the chatbots through this component which have many parts can be improved or adding some parts in the future if will be discovered that it will serve the DM. DM receives a user input from the NLU and produce the system responses at a concept level to the natural language generator (NLG). the response which the DM will choose is depending on the strategy that has been chosen. Strategies are related to maintaining conversational state and the ability to model the dialogue structure beyond that of a single utterance **[6]**. the strategies are rule-based, knowledge-based, retrieval based, and generative. The rule-based strategies are backstory, intent templates, and entity-based templates ordered by their priorities. Because rule based strategies encode human knowledge into the form of templates, they provide the most accurate responses. The system will adopt a template reply if input is recognized by one of these strategies. If there is no matching template for the input, the system can try to get an answer from a knowledge-base question answering (Q/A) engine **[5]**. Failing that, the input is handled by an ensemble of neural network models and information retrieval modules to create a general conversation output.

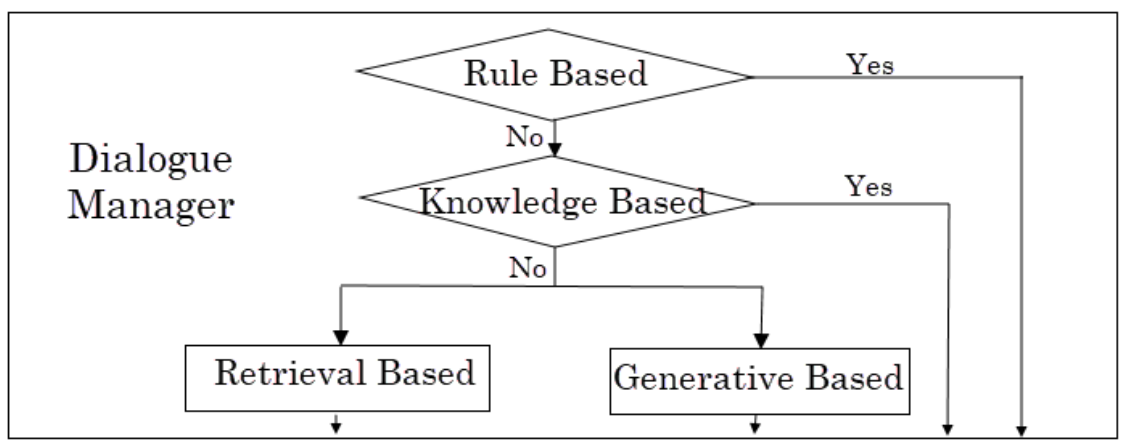

Figure 3: Dialogue Manager [Second Core Component]

1) *Online Information Retrieval:* This module can be used when the rule-based and the knowledge-based is fail to provide responses, it tries to provide more human-like, more concrete, and fresher responses compared to the rule-based templates and neural dialogue generation modules. The source of information for this module can be the most recent tweets provided by Twitter search API.*[14]* employed tweets as the source because they are usually short sentences closer to verbal language of most users compared to long written posts.

2) *Neural Network Reply Generation:* Generating sensitive context responses related to the conversation help in making the user feel more engaging. This should be used about taking into account the recent utterances. Deep Learning approach and other Machine Learning techniques can be applied to develop.*[15]*

D. *Natural Language Generator (Generator Reply)*

The last main core component of any chatbot. It receives a communicative act from the Dialogue Manager (DM) and generates a matching textual representation. There are two functions that the NLG must perform: content planning (Content Filter | Engagement Ranking) and language Generation using just Text| Speech using Text to Speech). After going through one or more of strategies of the DM, the pipeline proceeds to the reply generator. This generator firstly will apply a content filter out incoherent or questionable candidates. If there are more than one valid response, a ranking process is used to reorder candidate utterances firstly according to the priority and then according to the engagement ranking. Finally, the chatbot will send the selected utterance as a text in the final output format or sending it to the Text to Speech to generate the final output in Speech form. Simultaneously, all conversations are tracked in a history.

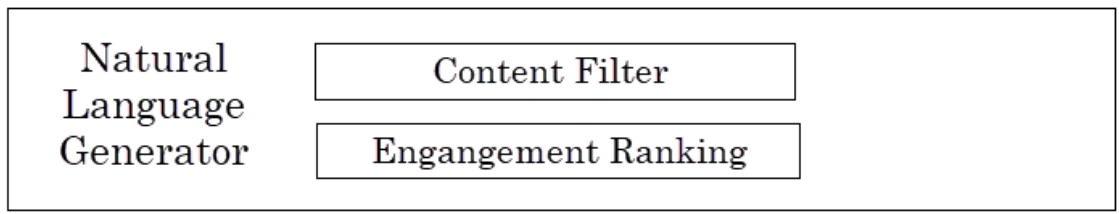

Figure 4: Natural Language Generator [Last Core Component]

## 4 CASE STUDY

KLM is a Dutch airline has an enormous number of flights around the world which was founded since 1919 so it was considered the world's oldest airline. KLM's owners thought about how they could improve their services for their clients. They found out that using the technology will be the key of success for this improvement. They are knowledge experts about having the information related to the area of study. They discovered that the social media is the most engagement way to the people. So they decided to ask the help from an AI experts about how they could benefit from the social media in the impact on the KLM airline services improvement.

A. *Discussion*

The main goal was to refining communication with their clients, making its client experience better by making it easier for people to talk to its agents via social media. KLM's owners studied their customers and knew that them spend a lot of time on social media like Facebook and Messenger. So it can be the entry point and this about enabling the clients from sending any inquiry and may be reservation message privately to KLM's social media page. Because of the huge number of messages, the AI experts come with some bots such as KLM bot which provide as many as services like immediate reply on messages that can reach to 60000 message, Try before you fly with augmented reality, Tip advisor, and booking your favorite seat with more comfortability, privacy and more. You can use this bot through Facebook Messenger, Telegram, Twitter, and more other Social Media.

*B. Evaluation*

They made a survey at the end of each Messenger conversation, about asking the client some questions like "How you can feel comfortability with our Messenger bot?", "Are you like to recommend this bot for a friend?", and so on. And the answers was options to choose from them scaled from [1-10].

This Messenger bot achieved a breakthrough through both meeting and exceeding the expectations. Since January 2017, the airline has achieved:

1) Increasing in the interaction with client through the Messenger reached to 40%.

2) The success of online boarding booking was 15% about this Messenger bot.

## 5 CONCLUSION

The discussion was around the chatbot or CA. Firstly we talked briefly about what is the CA and then turned into the main purpose of the paper about what is the chatbot system architecture and its main components and discussed each one in detail. NLU and it subs, DM and it subs, then finally the NLG or reply generator. And know how the DM is the most important part of any chatbot system because its components can be improved continuously or we can develop some alternative subs. If you interested in chatbots, You can use this paper as a reference either in developing chatbot system or working on improving the Dialog Manager components.

**REFERENCES**

[1] Heung-Yeung Shum, Xiaodong He, Di Li, Microsoft Corporation *"From Eliza to XiaoIce: Challenges and Opportunities with Social Chatbots"*, 2018

[2] Schuetzler, Ryan M.; Grimes, G. Mark; Giboney, Justin Scott; and Nunamaker, J ay F . Jr ." *The Influence of Conversational Agents on Socially Desirable Responding*" Proceedings of the 51st Hawaii International Conference on System Sciences The Influence of Conversational Agents on Socially, 2018

[3] Siddhant Rai, Akshayanand Raut, Akash Savaliya, Dr. Radha Shankarmani *"Darwin: Convolutional Neural Network based Intelligent Health Assistant"* Proceedings of the 2nd International conference on Electronics, Communication and Aerospace Technology (ICECA 2018), 2018

[4] ashwram, roprasad, ckhatri, anuvenk, raeferg, qqliu, jeffnunn, behnam, chengmc, nashish, kinr, kateblan, warticka, yipan, hasong, skj, ehwang, pettigru *"Conversational AI: The Science Behind the Alexa Prize"* 1st Proceedings of Alexa Prize (Alexa Prize 2017), 2018

[5] Boris Galitsky *Developing Enterprise Chatbots, 2019*

[6] Jurafsky D, Martin JH Speech and language processing (Pearson International), 2nd edn. Pearson/Prentice Hall, Upper Saddle River. ISBN 978-0-13-504196-3, 2009

[7] Skantze G ,Error handling in spoken dialogue systems-managing uncertainty, grounding and miscommunication. Doctoral thesis in Speech Communication. KTH Royal Institute of Technology. Stockholm, Sweden,2007

[8] Lee C, Jung S, Kim K, Lee D, Lee GG Recent approaches to dialog management for spoken dialog systems. Journal of Computing Science and Engineering 4(1):1–22,2010

[9] C. Khatri *et al.*, "Contextual Topic Modeling For Dialog Systems, "*IEEE Spoken Language Technology Workshop (SLT)*, Athens, Greece, 2018, pp. 892-899. doi: 10.1109/SLT.2018.8639552, 2018

[10] Intent Classification https://botfront.io/blog/how-intent-classification-works-in-nlu#teaching-semantics-to-a-machine.

[11] Haptik Open source chatbot NER :https://haptik.ai/tech/open-sourcing-chatbot-ner/ (accessed 12 January 2020

[12] Ferragina P, Scaiella U Tagme: on-the-fly annotation of short text fragments (by Wikipedia entities). In: Proceedings of the 19th ACM *international conference on information and knowledge management. ACM*, New York, pp 1625–1628, 2010

[13] Manning CD, Surdeanu M, Bauer J, Finkel J, Bethard SJ, McClosky The stanford CoreNLP natural language processing toolkit. Proceedings of 52nd Annual Meeting of the Association for Computational Linguistics: System Demonstrations, pp 55–60, Baltimore, Maryland USA, June 23–24,2014

[14] Liu H, Lin T, Sun H, Lin W, Chang C-W, Zhong T, Rudnicky A RubyStar: a non-task oriented mixture model dialog system. First Alexa Prize comptions proceedings,2017

[15] *A. Sordoni, M. Galley, M. Auli, C. Brockett, Y. Ji, M. Mitchell, J.-Y. Nie, J. Gao, B. Dolan. A Neural Network Approach to Context-Sensitive Generation of Conversational Responses. In Proc. of NAACL-HLT. Pages 196-205,2015*

**BIOGRAPHY**

Moataz Mohammed is a teaching assistant of Computer Science department, Ain Shams University, Cairo, Egypt. Master candidate student in NLP and DL fields, B.Sc. of Computer Science Dept., Faculty of Computer and Information Sciences, in 2017, Ain Shams University, Cairo, Egypt.

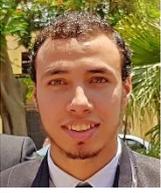

Prof. Mostafa M. Aref is a professor of Computer Science and Chairman of Computer Science department, Ain Shams University, Cairo, Egypt. Ph.D. of Engineering Science in System Theory and Engineering, June 1988, University of Toledo, Toledo, Ohio. M.Sc. of Computer Science, October 1983, University of Saskatchewan, Saskatoon, Sask. Canada. B.Sc. of Electrical Engineering - Computer and Automatic Control section, in June 1979, Electrical Engineering Dept., Ain Shams University, Cairo, EGYPT.

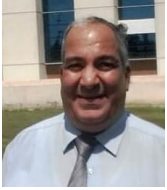

نبذه مختصرة - في واحدة من أكثر المجالات اهتماما في الفترة الأخيرة ألا وهي المحادثات الآلية حيث تستطيع ان تفتح البحث في أكثر من مجال في مجالات علوم الحاسب حيث انها تحتاج الى فهم كيف للحاسب ان يتعامل ويفهم لغة الإنسان وكيف يمكننا ان نجعل هذا الحاسب ان يطبق خصائص الذكاء الاصطناعي والتعلم العميق في إمكانية اخذ القرار والرد المناسب. في هذه الورقة البحثية سوف نناقش معا كيف يكون الهيكل البنائي لبناء أي محادثة آلية وسيتم هذا عن طريق مناقشة تفصيلية لكل مكون فيه.